\documentclass[lettersize,journal]{IEEEtran}
\usepackage{amsmath,amsfonts}
\usepackage{algorithmic}
\usepackage{algorithm}
\usepackage{array}
\usepackage[caption=false,font=normalsize,labelfont=sf,textfont=sf]{subfig}
\usepackage{textcomp}
\usepackage{stfloats}
\usepackage{url}
\usepackage{verbatim}
\usepackage{graphicx}
\usepackage{cite}
\usepackage{hyperref}
\usepackage[capitalize]{cleveref}
\crefname{section}{Sec.}{Secs.}
\Crefname{section}{Section}{Sections}
\Crefname{table}{Table}{Tables}
\crefname{table}{Tab.}{Tabs.}

\begin{document}

\title{BI-AVAN: Brain-inspired Adversarial Visual Attention Network}

\author{Heng Huang$^\ast$, Lin Zhao$^\ast$, Xintao Hu, Haixing Dai, Lu Zhang, Dajiang Zhu$^\dagger$, and Tianming Liu$^\dagger$
        % <-this % stops a space
\thanks{$\ast$ Heng Huang and Lin Zhao equally contribute to this study.}% <-this % stops a space
\thanks{$\dagger$ Co-corresponding authors.}
\thanks{Heng Huang and Xintao Hu are with the School of Automation, Northwestern Polytechnical University, Xi'an, China. (email: Huangheng05@163.com, xhu@nwpu.edu.cn)}
\thanks{Lin Zhao, Haixing Dai and Tianming Liu are with the Department of Computer Science, University of Georgia, Athens, GA, USA. (email: \{lin.zhao, hd54134, tliu\}@uga.edu)}
\thanks{Lu Zhang and Dajiang Zhu are with the Department of Computer Science and Engineering, The University of Texas at Arlington, Arlington, TX, USA. (email: lu.zhang2@mavs.uta.edu, dajiang.zhu@uta.edu)}
}

% The paper headers
%\markboth{Journal of \LaTeX\ Class Files,~Vol.~14, No.~8, August~2021}%
%{Shell \MakeLowercase{\textit{et al.}}: A Sample Article Using IEEEtran.cls for IEEE Journals}

%\IEEEpubid{0000--0000/00\$00.00~\copyright~2021 IEEE}
% Remember, if you use this you must call \IEEEpubidadjcol in the second
% column for its text to clear the IEEEpubid mark.

\maketitle

\begin{abstract}
Visual attention is a fundamental mechanism in the human brain, and it inspires the design of attention mechanisms in deep neural networks. However, most of the visual attention studies adopted eye-tracking data rather than the direct measurement of brain activity to characterize human visual attention. In addition, the adversarial relationship between the attention-related objects and attention-neglected background in the human visual system was not fully exploited. To bridge these gaps, we propose a novel brain-inspired adversarial visual attention network (BI-AVAN) to characterize human visual attention directly from functional brain activity. Our BI-AVAN model imitates the biased competition process between attention-related/neglected objects to identify and locate the visual objects in a movie frame the human brain focuses on in an unsupervised manner. We use independent eye-tracking data as ground truth for validation and experimental results show that our model achieves robust and promising results when inferring meaningful human visual attention and mapping the relationship between brain activities and visual stimuli. Our BI-AVAN model contributes to the emerging field of leveraging the brain’s functional architecture to inspire and guide the model design in artificial intelligence (AI), e.g., deep neural networks.
\end{abstract}

\begin{IEEEkeywords}
fMRI, visual attention, brain, brain-inspired AI
\end{IEEEkeywords}

\section{Introduction}
\IEEEPARstart{V}{isual} attention refers to the capability of selectively focusing on part of a visual scene rather than the entirety, given the limited processing capacity of the human visual system \cite{ungerleider2000mechanisms}. Inspired by this fundamental biological process of the brain, a large group of deep learning studies successfully integrated attention mechanism into their deep neural networks for improving the performance and the interpretability \cite{hassabis2017neuroscience,vaswani2017attention,qiuxia2020understanding}. For example, in the computer vision (CV) field, a lightweight attention module has been introduced into convolutional neural networks (CNNs) and demonstrated consistent improvements on both image classification and object detection tasks \cite{woo2018cbam}. Also, those attention based neural networks have been employed to predict and study the human visual attention \cite{liu2015predicting,wang2017deep,qiuxia2020understanding}. In general, bridging the gap between brain science and artificial intelligence, e.g., incorporating the attention mechanism into deep learning can not only inspire and guide the design of neural networks with better performance/interpretability but also facilitate the understanding of our human brain.

\begin{figure}[t]
  \centering
  %\fbox{\rule{0pt}{2in} \rule{0.9\linewidth}{0pt}}
   \includegraphics[width=1.0\linewidth]{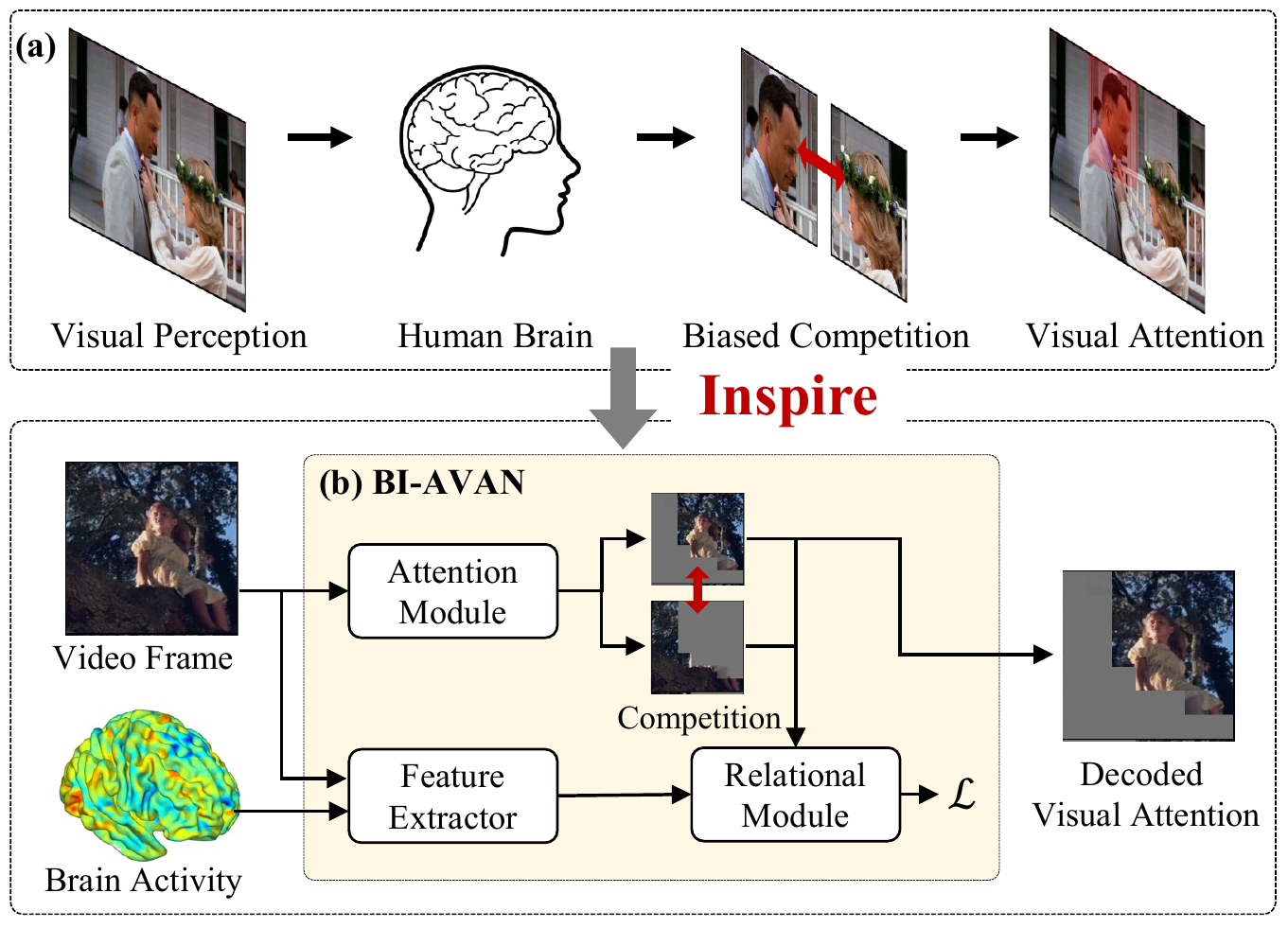}

   \caption{(a) An illustration of biased visual competition in human brain. (b) Overview of the proposed BI-AVAN model. Inspired by the biased competition in human brain, the attention module outputs the attention-related/neglected contents for visual competition to decode the human visual attention from brain activity. }
   \label{fig:figure1}
\end{figure}

In previous visual attention studies, eye-tracking based technology has been the dominant method to characterize human visual attention\cite{liu2011visual,liu2015predicting,wang2017deep,sood2020interpreting}. Researchers have demonstrated the close relationship between eye movements and visual attention \cite{sheliga1994orienting,liu2016makes}. However, as an indirect way to study human visual attention, e.g., by linking the locations of eye gaze and brain attention, eye-tracking based methods may be limited in comprehensively and robustly characterizing human visual attention. For instance, some studies have reported that eye-tracking based methods may ignore the complex selective process of human brains including visual recognition, memory retrieving, and social perception \cite{ungerleider2000mechanisms}. Therefore, some studies turned to a more natural way by using functional magnetic resonance imaging (fMRI) to directly study visual attention from brain activity \cite{kanwisher2000visual,hu2015sparsity,parhizi2018decoding}, and their results indicate the importance and superiority of fMRI in studying visual attention. But these studies only focus on the neuroscientific findings of visual attention, such as which brain regions (mainly visual cortex) are related to specific visual attention tasks (visual stimulus). Using functional brain activities to characterize and represent attention remains largely unexplored. Meanwhile, many brain science studies suggested that human visual attention has a biased competition (the top-down mechanisms of biased competition theory) \cite{duncan1997competitive,beck2009top}. That is, the capacity of an individual’s visual system is limited, and visual objects have to compete for the limited brain resources (\cref{fig:figure1}(a)). Hence, the attention-related objects and the neglected background always have an adversarial relationship: at the neural level, whenever there is a widespread maintenance of the attention related object’s representation, there is a widespread suppression response to neglected background objects at the same time \cite{duncan1997competitive}. Unfortunately, this adversarial relationship has not been fully exploited, neither in neural networks design nor in characterizing the human visual attention using deep learning methods.

To bridge these gaps, in this paper, we developed a Brain-inspired Adversarial Visual Attention Network (BI-AVAN) to characterize and decode the visual attention in a more real-world and complex scene (movie watching) with fMRI-derived brain activities. Specifically, in the visual attention decoding process, our BI-AVAN model imitates the biased competition process between attention-related and neglected objects (\cref{fig:figure1}(b)). To do so, we introduced an attention module to divide the outer environment (e.g., an image or a frame of the movie) into attention-related and neglected parts in an adversarial manner, in which each visual object in the image can only belongs to one of them. A relational module is then employed to maximize/minimize the relation between the attention-related/neglected parts and the brain activities. The rationale behind our BI-AVAN model design is that attention-related parts gain dominance in brain activities while neglected parts are suppressed in the brain. Thus, we can use the brain activities to guide the BI-AVAN model in the training stage. In the inference stage the attention module can locate the attention-related objects which are more coherent with brain activities. We adopted an fMRI dataset with simultaneous eye-tracking data to evaluate the performance of proposed BI-AVAN model. The experimental results demonstrate that the BI-AVAN model can effectively and robustly decode group-wise and individual-specific human visual attention. The brain networks identified from our BI-AVAN model are meaningful and have a close relationship with the biased competition in the human brain. The objects of interests in human visual attention are also analyzed based on the inferred attention-related content. Overall, our BI-AVAN model provides novel insights on the computational aspects of the visual attention mechanism and contributes to the emerging field of brain-inspired AI.

\section{Methods}

\subsection{Overview}

Our proposed BI-AVAN model is an imitation of the biased competition process in human visual attention. \Cref{fig:figure2} illustrates the major modules of the BI-AVAN model, which consists of an attention module for locating the visual attention (\cref{fig:figure2}(a)), a feature encoding module for extracting both image and brain activity features (\cref{fig:figure2}(b)), and a relational module for discriminating attention-related and attention-neglected content (\cref{fig:figure2}(c)).

\subsection{Attention Module with Adversarial Learning}
\label{sec:attention_module}
The attention module is designed to imitate the biased competition process in human visual attention for characterizing the attention-related and attention-neglected parts. In order to maintain the adversarial relationship between these two parts in the attention module, the pixels of the input image are enforced to belong to only one of the two parts: mathematically, we use $F(x)$ to map pixel $x$ to a possibility value $\alpha$ which denotes the probability of the pixel $x$ belonging to the attention-related part. The possibility of pixel $x$ belonging to the attention-neglect part is then represented as $1-\alpha$ to maintain the adversarial relationship. In this section, we illustrate the architecture of residual network in the attention module of BI-AVAN.

\begin{figure*}[t]
  \centering
  %\fbox{\rule{0pt}{2in} \rule{0.9\linewidth}{0pt}}
   \includegraphics[width=1.0\linewidth]{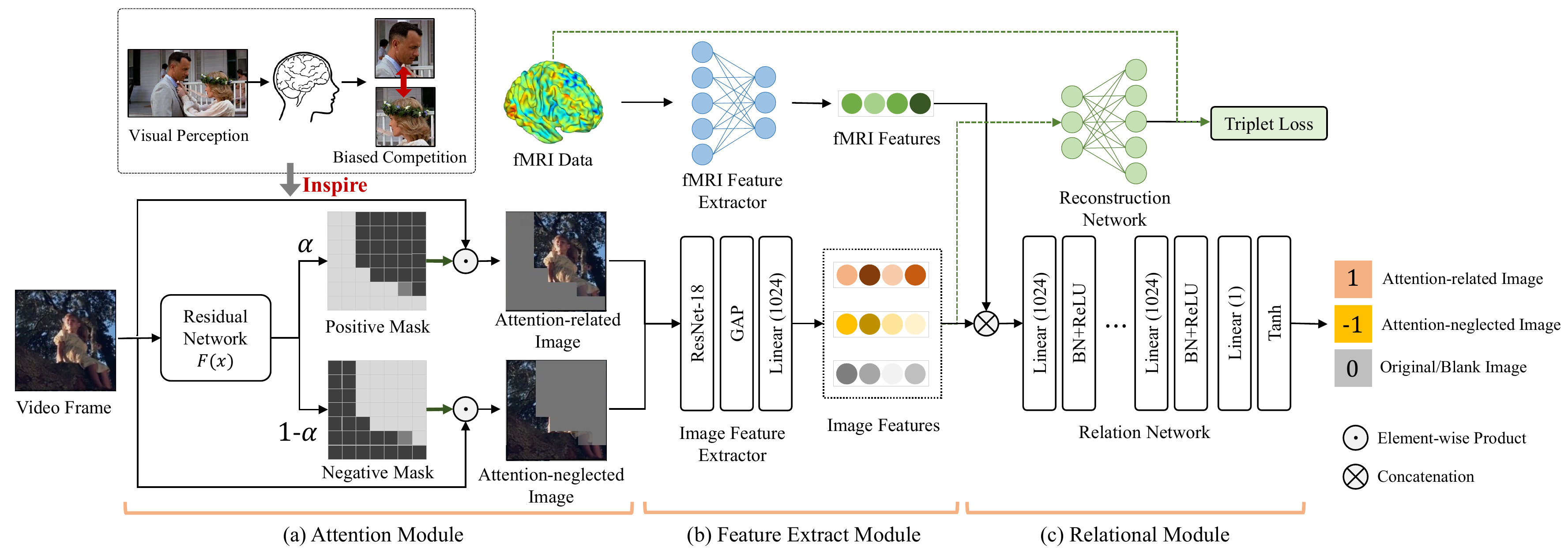}

   \caption{The proposed BI-AVAN framework. (a) shows the overall computational pipeline of the attention module. With an input image, the residual network generates two possibility matrices (denote as $\alpha$ and $1-\alpha$, respectively). Attention-related and attention-neglected content are obtained by dot product of the original image with the upsampled $\alpha$ and $1-\alpha$. (b) illustrates the feature extractor module which consists of an fMRI feature extractor and an image feature extractor. The concatenation of image features and fMRI features are input into (c) relational module to maximize the distance between attention-related/neglected contents.}
   \label{fig:figure2}
\end{figure*}

%In this section, we illustrate the architecture of residual network in the attention module of BI-AVAN. 

\begin{figure}[ht]
  \centering
   \includegraphics[width=0.75\linewidth]{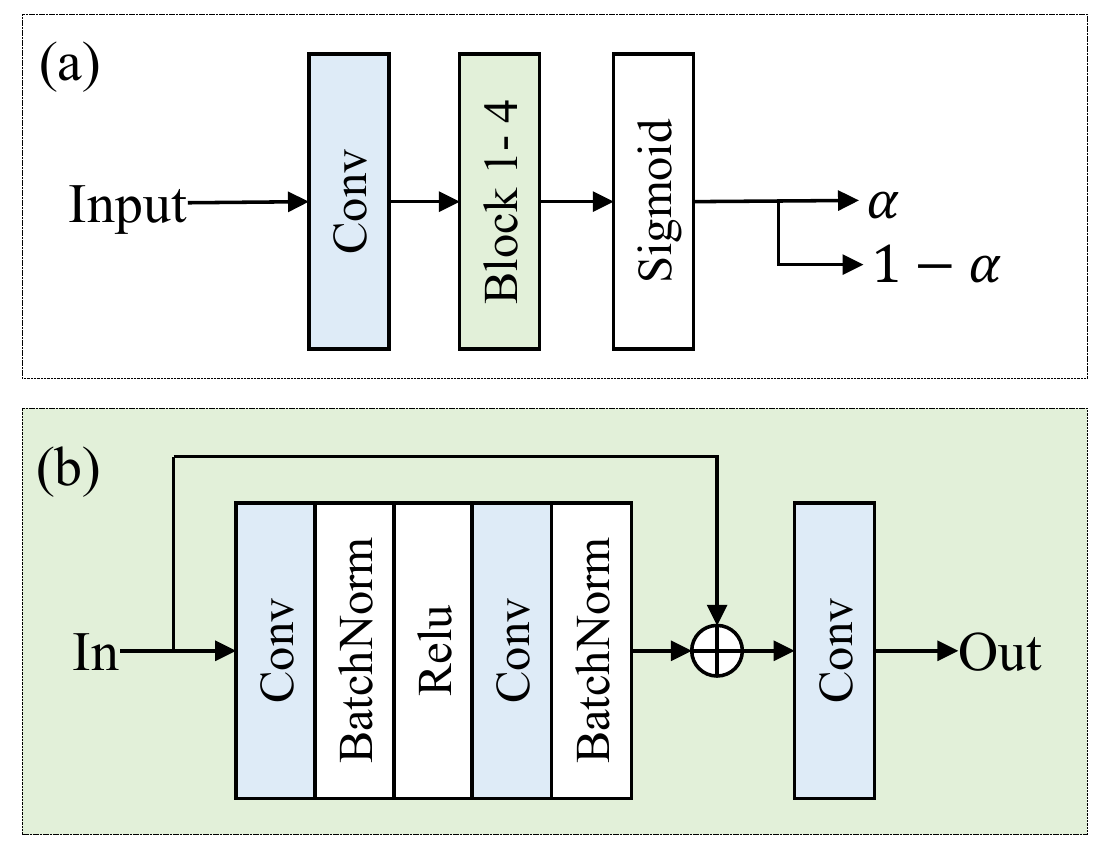}

   \caption{The architecture of the residual network in the attention module. (a) The residual network contains one 2D convolution layer and four residual building blocks, and the structure of which is shown in (b).}
   \label{fig:figureS1}
\end{figure}

In this study, we used a residual neural network (ResNet-18) \cite{he2016deep} as the core component of our attention module to learn the function F(x). We removed the fully connected layer in ResNet-18 and modified the last convolution layer to produce a probability matrix $\alpha$ which represents the current probability of attention-related content (\cref{fig:figure2}(a)). $1-\alpha$ represents the probability of attention-neglected content in the image. Specifically, as shown in \cref{fig:figureS1}(a), the residual network contains one 2D convolution layer and four residual building blocks (\cref{fig:figureS1}(b)) followed by a Sigmoid activation function. For an input image, the residual network will process it layer by layer and output the probabilistic matrices $\alpha$ and $1-\alpha$. The residual building blocks input the images/feature maps into a 2D convolution layer followed by Batch Normalization (BN) and Relu activation function. Then, the output of another convolution layer with BN is concatenated with the input and passed into a final convolution layer to obtain the final output. Both $\alpha$ and $1-\alpha$ have the size of $7\times7$, which can be up-sampled to the original image size as two probability masks. The segmentation of attention-related/neglected parts are obtained by the dot product of the input image with the two probability masks, respectively. After segmentation, the resulted attention-related/neglected images will be the input of the feature encoding module (\cref{sec:encoding_module}) for feature extraction.

\subsection{Feature Encoding Module}
\label{sec:encoding_module}

\begin{figure}[ht]
  \centering
   \includegraphics[width=1.0\linewidth]{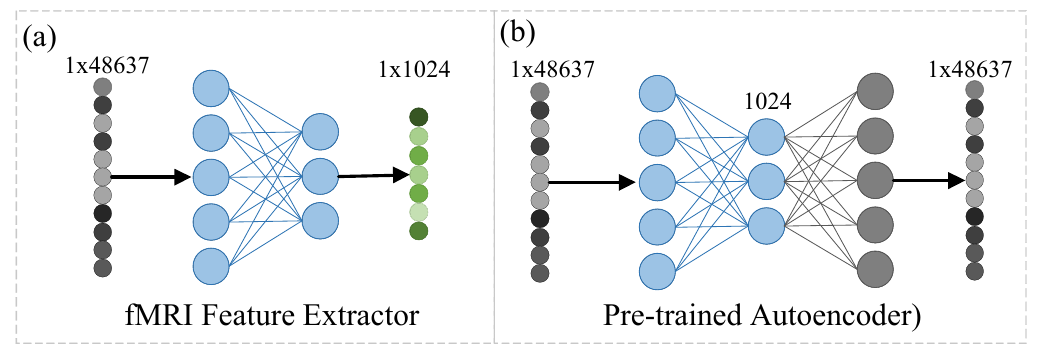}

   \caption{(a) The fMRI feature extractor is a regular feedforward neural network. Before the model training, it will be initialized with the encoder's weights in a pre-trained autoencoder (b).}
   \label{fig:figureS2}
\end{figure}

The feature encoding module consists of an image feature extractor for attention-related/neglected images and an fMRI feature extractor. Both image and fMRI data will be encoded as feature vectors with sizes of $1\times1024$. For image feature extractor, we used ResNet-18 and replaced the last average pooling layer with a global average layer. The fully connected layer was correspondingly modified with 1024 units and tanh activation function. For fMRI feature extractor, we used a single fully connected layer to conduct a linear decomposition of fMRI data. A L1 regularization (with a penalty coefficient 5e-6) was applied to introduce the sparsity of the decomposition for the convenience of brain activity pattern visualization \cite{lv2015sparse}. It is noted that we adopted a pre-trained fMRI autoencoder to initialize the fMRI feature extractor. Specifically, before starting the training of the whole BI-AVAN model, we initialize the fMRI feature extractor with the encoder's weights in a pre-trained fMRI autoencoder (\cref{fig:figureS2}). Considering that fMRI data is very high-dimensional, if we try to directly optimize the fMRI feature extractor, the number of parameters will be $48637\times1024$. Notably, the acquisition of fMRI is very expensive and time-consuming, and we only have 53,985 fMRI volumes in this study. So, such initialization can help reduce the overfitting and make model converge much faster. In our experiments, we found that by initializing the fMRI feature extractor with a pre-trained fMRI autoencoder, the model's performance can be significantly improved.

\subsection{Relational Module}
\label{sec:relation_module}
The relational module is the core component in our BI-AVAN model. Here, we assume the brain activities should show different patterns when focusing on the attention-related and attention-neglected content in the movie. Therefore, the goal of the relational module is to discriminate the attention-related content from attention-neglected content by maximizing/minimizing the relationship between the brain activities (fMRI features) and the potential attention-related/neglected pixels in the images. Here, we formulate the relational module as a neural network $f_{rel}(\cdot)$. \cref{fig:figure2}(c) is an illustration of the relational module consisting of multiple fully connected layers followed by batch normalization. Specifically, the feature vectors ($1\times1024$) of the attention-related/neglected images ($v_a$/$v_n$) and fMRI data ($v_f$) are concatenated as a single vector ($1\times2048$). The relational module takes it as input and outputs a scalar, ranging from -1 (for attention-neglected image) to 1 (for attention-related image). In addition, we introduced two additional regularization terms in the relational module. The first regularization term makes the concatenation of the fMRI feature vector and the original image feature vector towards to 0 (original image = attention related contents + neglected contents). This regularization term will prevent the BI-AVAN model from identifying the entire image as attention related or neglected content. The second regularization term makes the concatenation of feature vectors from a blank image and fMRI data towards 0, which can effectively exclude the influence of fade in (fade out) movie frames. 
The training objective of the relational module is then formulated in \cref{eq:eq1}:
\begin{align}
\mathcal{L}_{rel} = & (1-f_{rel}(v_{af}))^2+(-1-f_{rel}(v_{nf}))^2 \nonumber\\ 
&+f_{rel}(v_{anf})^2+f_{rel}(v_{bf})^2
\label{eq:eq1}
\end{align}%
where $v_{af}$, $v_{nf}$ and $v_{bf}$ represent the concatenation of fMRI feature vector, the feature vector of attention-related, attention-neglected and blank images, respectively. $v_{anf}$ is the concatenation of fMRI feature vector and the vector of original image (equals to $v_{a}+v_{n}$).
We also designed a triplet loss term to exclude some randomness of BI-AVAN model by introducing an fMRI reconstruction network $f_{rec}(\cdot)$ which is consisted of a fully connected layer followed by batch normalization. We use the feature vector of attention-related/neglected images to reconstruct the original fMRI data. The rationale behind this triplet loss is that the attention-related images should have a stronger relation to the brain activities and thus should have lower reconstruction error compared to neglected ones. By introducing the triplet loss, the order of attention related contents (neglected contents) will be fixed. The equation of triplet loss is shown in \cref{eq:eq2}:
\begin{equation}
\mathcal{L}_{trip}= max\{d(s,f_{rec}(v_a))-d(s,f_{rec}(v_n))+m,0\}
\label{eq:eq2}
\end{equation}%
where the $d$ is Euclidean distance function, $s$ is the original fMRI data. We used a very small margin value ($m=0.1$) to avoid the reconstruction network from dominating the training of relational module. 
By combining the \cref{eq:eq1} and \cref{eq:eq2}, the final loss function of BI-AVAN model is shown in \cref{eq:eq3}:
\begin{equation}
\mathcal{L} = \mathcal{L}_{rel}+\mathcal{L}_{trip}
\label{eq:eq3}
\end{equation}%

In general, the training objective of our BI-AVAN model is to minimize the loss function \cref{eq:eq3}. In this study, we trained our model in a GeForce GTX 1080Ti graphics card with Adam optimizer, it takes us 51.5 hours to train the entire model.

\subsection{Individual-specific Attention Related Content}
\begin{figure}[t]
  \centering
  %\fbox{\rule{0pt}{2in} \rule{0.9\linewidth}{0pt}}
   \includegraphics[width=1.0\linewidth]{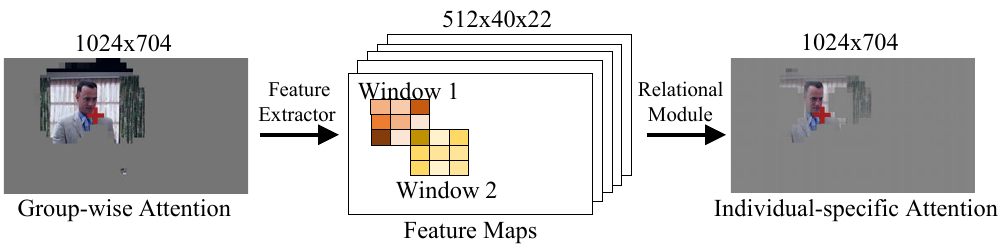}

   \caption{The computational pipeline of individual-specific attention. Each element on relational map represents the corresponding image blocks’ relation to brain activities. In this figure, input image is center cropped for illustration purpose.}
   \label{fig:figure3}
\end{figure}

It is worth noting that our attention module does not combine any individual brain activities or eye-tracking information to infer the visual attention, which means the derived attention related content is an inference of group interest. To obtain the individual-specific attention, first, we used the image feature extractor (the global average and dense layer are temporarily removed) to encode it as feature maps. As shown in \cref{fig:figure3}, for each movie frame we can obtain 512 feature maps (feature map size is 40x22), and each location in the feature maps corresponds to a 32x32 image block in the original movie frames (the size of receptive field is 32x32). Then, for each location in the feature maps, we applied a sliding window (window size is 3x3) to achieve a corresponding image codes vector (with the global average layer and dense layer). The code vector is then concatenated with individual’s fMRI code vector to generate a relational value from the relational module. The relational values represent the correlation strength between image content and brain activities. The final individual-specific attention is obtained by dot product of the original image with individual’s relational map (the relational map is up-sampled to the same size of the original image).

\section{Experimental Results}

\subsection{Dataset}
In this study we used a public movie dataset with simultaneous fMRI and eye-tracking data (Forrest Gump dataset, \cite{hanke2016studyforrest}). The fMRI and eye-tracking data were recorded during watching a movie (2 hours of Forrest Gump). The dataset contains 15 subjects with 53,985 fMRI volumes (each subject has 3,599 fMRI volumes and each fMRI volume has 48,637 valid voxels). To preprocess the fMRI data, we used the standard and widely used fMRIprep preprocessing pipeline \cite{esteban2019fmriprep}. The preprocessing steps include: skull stripping, motion correction, slice time correlation, registration to standard the MNI space and global drift removal. More details are referred to \cite{esteban2019fmriprep}. After the standard preprocessing, we normalized voxels’ signal with zero mean and standard deviation, and then applied a pre-defined brain mask to extract valid fMRI signals from these preprocessed 3D fMRI volumes. The extracted signals are flattened as 1D vectors for model training and validation. In our experiment, we have 15 subjects in total. Each subject has 3,599 fMRI volumes and each fMRI volume contains 48,637 valid fMRI time series signals.

The movie used in this study was encoded as H.264 video with a resolution of $1280\times720$ pixels (25fps). It contains 179,926 movie frames in total, and the viewing distance is 63cm which ensures that the participants can see the full screen \cite{hanke2016studyforrest}. For eye-tracking data preprocessing, we excluded the off-screen eye movements as well as the slow eye blinks (blink more than 300 milliseconds). Median filter with a window size of 40 was applied to reduce the noise in the eye-tracking data. According to the biased competition theory, the competition of visual objects only happens within the human visual field, which means for a large movie screen, only the objects around eye gaze point competes for the human attention. Thus, to obtain more accurate training data for every movie frame, we cropped the original movie frame around the eye-tracking points to a size of 224x224. By doing so, we implicitly introduced an assumption that all the cropped image contains both attention-related content and attention-neglected content. However, this assumption could be inaccurate considering the selective process of human brain. It is possible that the cropped image does not contain any attention-related content if it is not selected by the brain. Thus, to exclude the uncertainty, we used a regularization term in the relational module to force all original images to have zero relation values. With this regularization term, the BI-AVAN model is forced to not build any relationships between the entire movie frame and the brain activities.

The eye movements of subjects were recorded at the frequency of 1000Hz. It is noted that the frequency for fMRI (0.5Hz), eye-tracking (1000Hz) and movie (25Hz) data are different. In order to eliminate such frequency difference and build the correspondence among different data sources, we downsampled the eye-tracking data to the same frequency as movie and up-sampled the fMRI data by interpolation.

In this study, we used 70\% of the samples for model training and the rest 30\% for testing. Unless we specifically mentioned, all the experimental results are based on testing data. 

\subsection{Evaluation of Group-wise Visual Attention}

\begin{figure}[t]
  \centering
  %\fbox{\rule{0pt}{2in} \rule{0.9\linewidth}{0pt}}
   \includegraphics[width=1.0\linewidth]{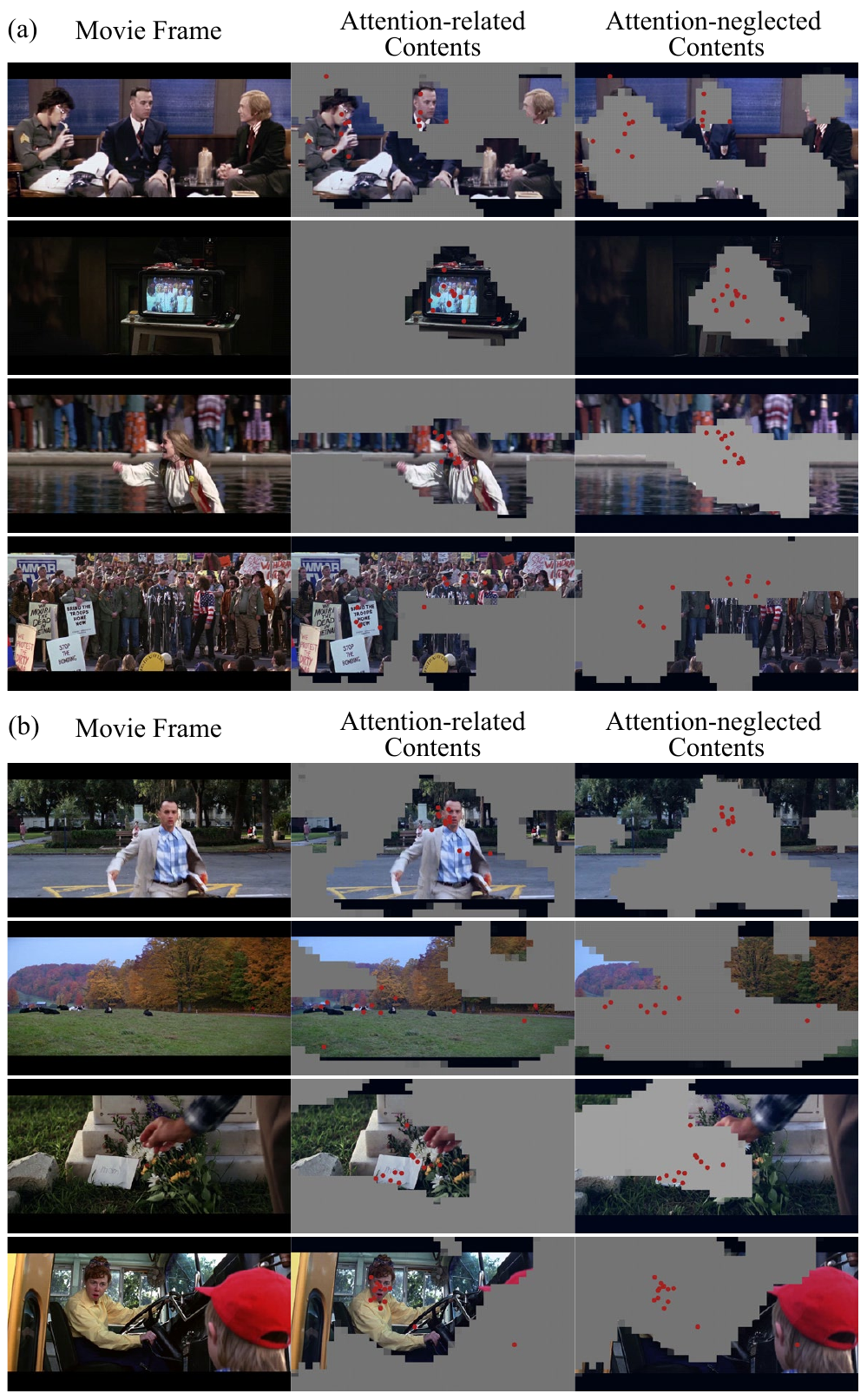}

   \caption{The group-wise attention-related content and attention-neglected content of BI-AVAN model from 4 randomly selected movie frames. Eye-tracking points are marked as red dots. (a) The segmentation results on training data; (b) The segmentation results on testing data.}
   \label{fig:figure4}
\end{figure}

We first evaluated the performance of BI-AVAN model in generating the attention-related/neglected content. Since the attention-related content we directly obtained from the attention module is an inference from group interest, it contains all possible objects that the participants might be interested in. \cref{fig:figure4} displays some random-selected results from training data (\cref{fig:figure4}(a)) and testing data (\cref{fig:figure4}(b)). More examples can be found in appendix. An interesting observation is that our attention module performs a semantic segmentation on image when identifying attention-related/neglected content: for most of the movie frames, the backgrounds (such as sky, ground, buildings, trees) are considered as attention-neglected content while other objects (such as people, poster, television, letters) are considered as attention-related content. We further compared the generated attention-related content to the results of eye-tracking. The corresponding eye-tracking points (the locations of the eye gaze) are marked as red dots in \cref{fig:figure4}. On both training and testing data, we found that most of the eye-tracking points are located within the attention-related regions instead of neglected regions. To quantitatively evaluate the performance of our attention module, we calculated the hit rates (hit rate equal to the ratio of eye-tracking points that are located within the attention-related region among all the eye-tracking points). For all 15 subjects, we have 1,586,459 eye-tracking points on the training data and 679,911 eye-tracking points on the testing data. We achieved 0.7793 and 0.5951 hit rate on training and testing data, respectively. The relatively high hit rates suggest that the attention-related part in our attention module has relatively higher possibility to draw participants’ attention than the neglected part.

We also calculated the averaged values of the positive/negative/regularization terms in the relational module, as summarized in \cref{tab:table1}. In general, the outputs of relational module indicate that the distance between attention-related and attention-neglected components, as expected, have been successfully maximized. Meanwhile, we observed the output values on training data are closer to targets than those on testing data, which suggests that over-fitting may exist due to the lack of fMRI data, given 53,985 fMRI volumes versus 179,926 movie frames. However, in our experiments, we found the over-fitting can be alleviated by initializing the parameters of fMRI feature extractor using a pre-trained fMRI autoencoder.

\subsection{Individual-specific Visual Attention and Eye Movements}

\begin{table}
\centering
\begin{tabular}{cccc}
\hline
  & Positive & Negative & Regularization\\
\hline
Target & 1  & -1 & 0 \\
Training dataset & 0.91  & -0.94 & 0.023\\
Testing dataset & 0.86  & -0.83 & 0.12\\

\hline
\end{tabular}
\caption{The average output of relational module on training data and testing data}
\label{tab:table1}
\end{table}

\begin{figure}[t]
  \centering
  %\fbox{\rule{0pt}{2in} \rule{0.9\linewidth}{0pt}}
   \includegraphics[width=1.0\linewidth]{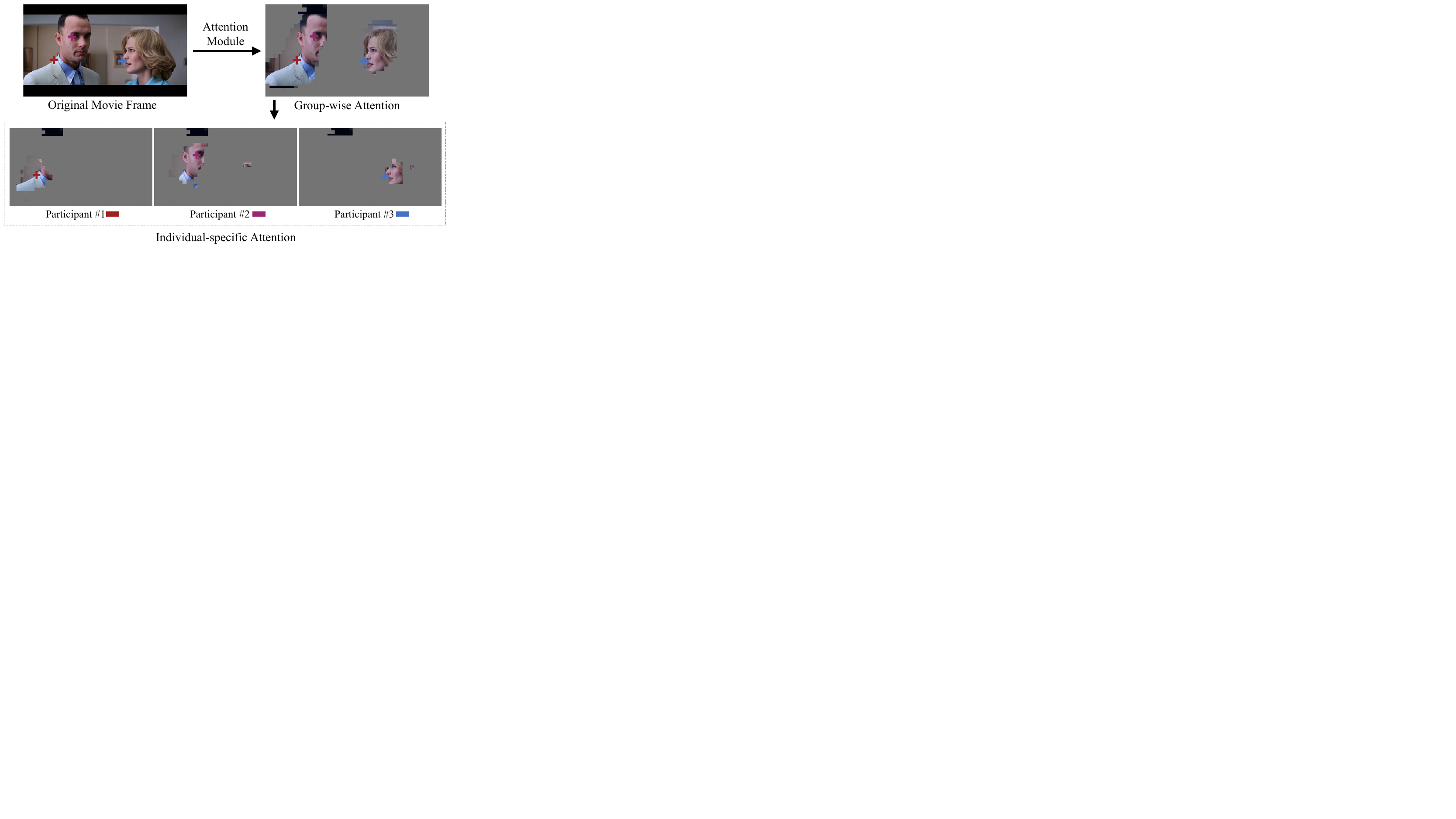}

   \caption{An example of group-wise attention and individual-specific attention. The eye-tracking points of different subjects are marked with different colors.}
   \label{fig:figure5}
\end{figure}

\begin{figure}[t]
  \centering
  %\fbox{\rule{0pt}{2in} \rule{0.9\linewidth}{0pt}}
   \includegraphics[width=1.0\linewidth]{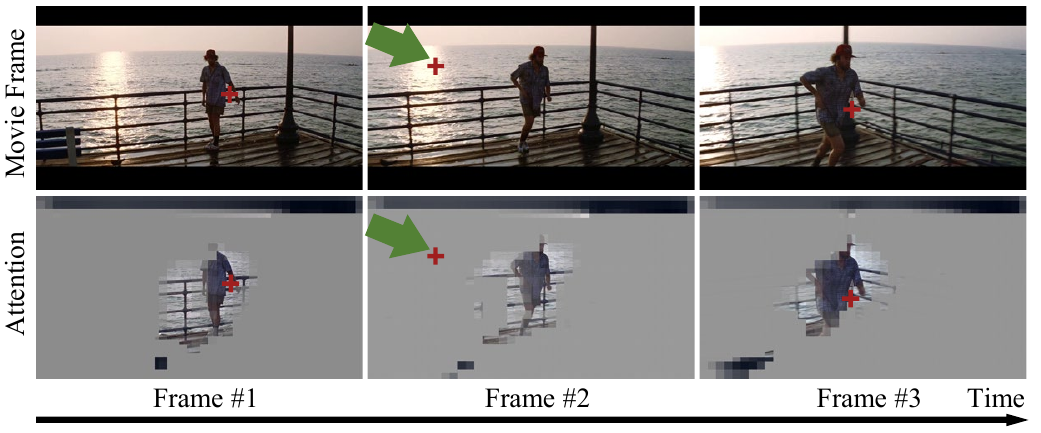}

   \caption{The illustration of human visual search. The images on top row are original movie frames, while those on the bottom row are the individual-specific attention. The eye-tracking points are highlighted as red cross-hair. The visual search happened in frame \#2, where the individual-specific attention and eye-tracking point are mismatched.}
   \label{fig:figure6}
\end{figure}

In this section, we focus on individual-specific attention-related content and discuss its relationship with eye movements. \cref{fig:figure5} shows an example of the obtained group-wise attention map as well as three individual-specific attention maps (participant \#1-\#3). Their corresponding eye-tracking points are highlighted in red, purple and blue color, respectively. More examples can be found in appendix. We can see that the individual-specific attention maps tend to be subsets of group-wise attention. The eye-tracking points are properly located in the individual-specific content, suggesting close relationship between individual attention and brain activities. Like the group-wise attention, we use the eye-tracking data to verify the individual results. Our model achieves 0.4189 (training data) and 0.3685 (testing data) hit rate on individual-specific attention. To investigate the reason of hit rate decrease, we visualized the individual-specific attention-related content maps and compared them to their corresponding eye-tracking points. We found that the decrease of hit rate is mainly caused by the random visual search of participants. \cref{fig:figure6} illustrates a random case with three frames in a movie clip. We can see that during  the movie watching, the participant quickly moved his/her eye gaze from Forrest to the sunset in Frame \#2 and then moved back to Forrest in Frame \#3. Although the eye movement happened in a real situation, our model will not capture it (highlight with green arrow in \cref{fig:figure6}) due to the sampling frequency of fMRI. The quick eye gaze movements here are usually related to human visual search, which is inevitable, since it is human instinct to scan visual environment for objects \cite{horowitz1998visual}. 

\subsection{Delay of Hemodynamic Response}

\begin{figure}[ht]
  \centering
   \includegraphics[width=1.0\linewidth]{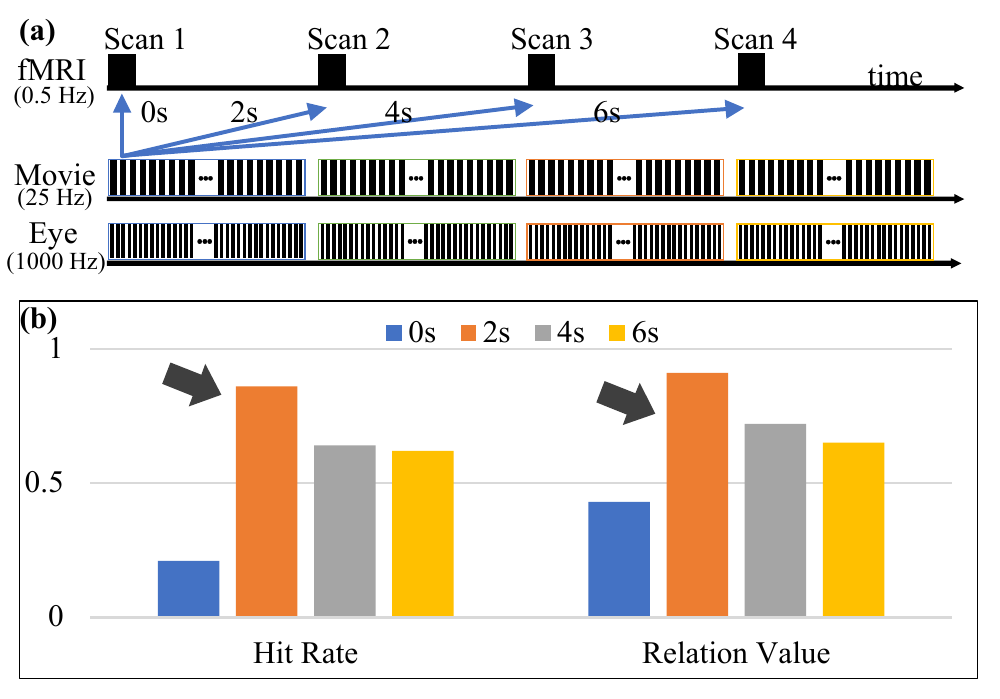}

   \caption{(a) An illustration of the frequency difference between different data sources. The black boxes represent the presence of samples. The movie frames and eye-tracking points between each two fMRI scans are highlighted in different colors. (b) The influence of different delay times with respect to the model's performance.}
   \label{fig:figureS3}
\end{figure}

It is noted that the brain activities measured by fMRI always lag behind the events (stimulus) due to the delay of hemodynamic response (HDR) \cite{aguirre1998variability}. Therefore, we need to align the fMRI with the movie to eliminate the delay caused by HDR. We did several experiments to find the delay time for our case by assuming the delay time is 0s (no lag), 2s, 4s and 6s, respectively. For each assumption, we trained a corresponding BI-AVAN model and evaluated their performances according to their attention locating abilities (hit rate with eye-tracking points) as well as the relation value (output value) of attention-related components in the relational module. The comparison results are shown in \cref{fig:figureS3}. From these experiments, we observed the best results when the delay time is 2s (in terms of the highest hit rate and the highest relation value, highlighted with black arrow) and the worst performance when delay time is 0s. In \cref{fig:figureS3}, it is interesting to see relatively good results when the delay time is 4s and 6s. This might be related to the memory of the human brain \cite{nichols2006working}. Our experiment results are also consistent with previous fMRI study \cite{deyoe1994functional} which suggested that the fMRI response evoked by visual stimuli delayed 1-2s and reached 90\% of peak in 5s. All experimental results in this study are based on the 2s delay time. However, we should point out that although we achieved reasonable results in this study, 2s delay time might not always be the optimal delay time for natural stimulus studies. We are not able to continue narrowing down the range of the optimal delay time (a value between 0-2s) due to the physical limitations of fMRI. In the future, the results could be improved by utilize a higher temporal resolution fMRI scanner. 

\subsection{Brain Networks Learned by BI-AVAN Model}

In this subsection, we evaluate if the BI-AVAN model learns meaningful brain networks from raw fMRI data for visual attention decoding. In BI-AVAN, the fMRI feature extractor is responsible for encoding the input fMRI signal to a feature vector. After the training process, the weight matrix of fMRI feature extractor contains the patterns learned from raw fMRI signals and each pattern can be interpreted as a specific brain network \cite{lv2015sparse}. The weight matrix has the size of $1024\times48637$, thus each row of the matrix represents a brain network. To visualize these brain networks, the values in each row are mapped back to the brain volume space and results in 1024 brain networks in total. The original fMRI data can be represented as a linear combination of these 1024 brain networks \cite{lv2015sparse}.
\begin{figure}[t]
  \centering
  %\fbox{\rule{0pt}{2in} \rule{0.9\linewidth}{0pt}}
   \includegraphics[width=1.0\linewidth]{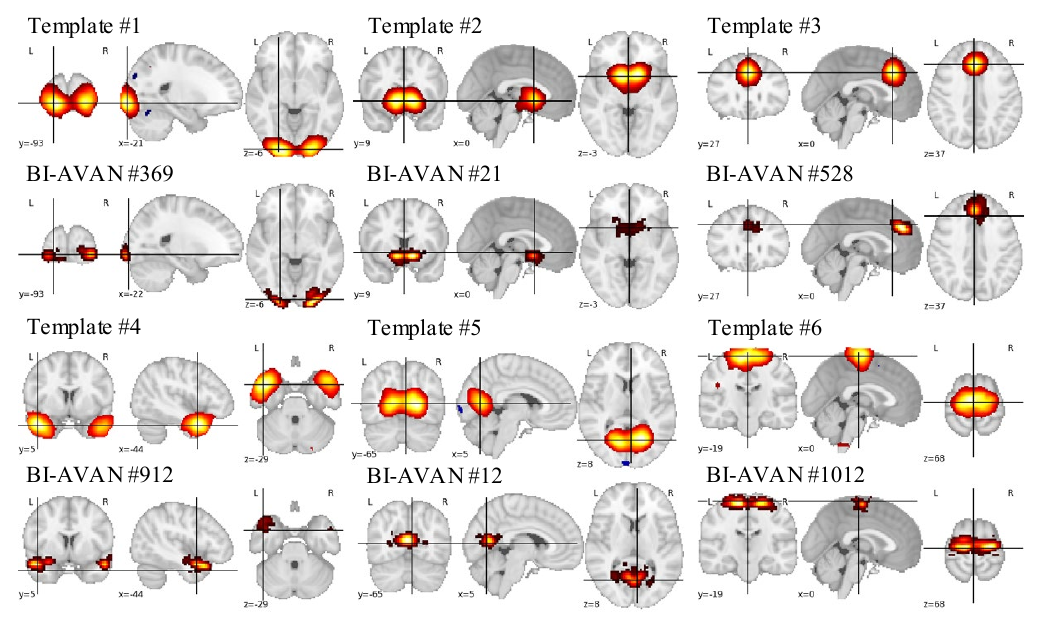}

   \caption{The comparison between the learned brain networks from BI-AVAN model and brain network templates shown in the same orthogonal slices with the same threshold value (3.0).}
   \label{fig:figure7}
\end{figure}

\begin{figure}[t]
  \centering
  %\fbox{\rule{0pt}{2in} \rule{0.9\linewidth}{0pt}}
   \includegraphics[width=1.0\linewidth]{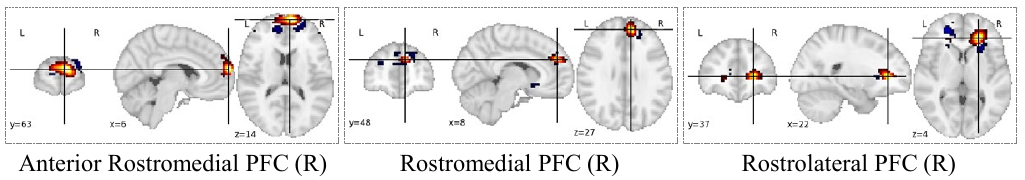}

   \caption{The rostral prefrontal cortex (rostral PFC) obtained from BI-AVAN model as three separate subareas. Only the right hemisphere is shown.}
   \label{fig:figure8}
\end{figure}

To verify if the learned brain networks are neurologically meaningful, we compared them with the widely used network templates \cite{smith2009correspondence} which include typical functional networks in human brain. Almost all the brain network templates can be found in our results (\cref{fig:figure7}). It is also inspiring that we found several brain networks which are highly related to the biasing attention of human brain. Previous fMRI studies \cite{henseler2011gateway} demonstrated there is a control system located in rostral prefrontal cortex (rostral PFC). The rostral PFC acts as a gateway of human attention and plays a key role in balancing attentional orienting to external and internal information. In this study, the rostral PFC identified from BI-AVAN model are shown as three separate sub-regions in \cref{fig:figure8}. It is consistent with \cite{henseler2011gateway} where they demonstrated the functional segregation exists between medial part and lateral part of rostral PFC. Specifically, the medial part (Rostromedial PFC) is responsible for processing external information by interacting with other parts of the brain, the lateral part (Rostrolateral PFC) processes the internally represented information and the anterior rostromedial PFC has relation to attentional biases generation. 

\subsection{Objects of Interests in Visual Attention}
\begin{figure}[t]
  \centering
  %\fbox{\rule{0pt}{2in} \rule{0.9\linewidth}{0pt}}
   \includegraphics[width=1.0\linewidth]{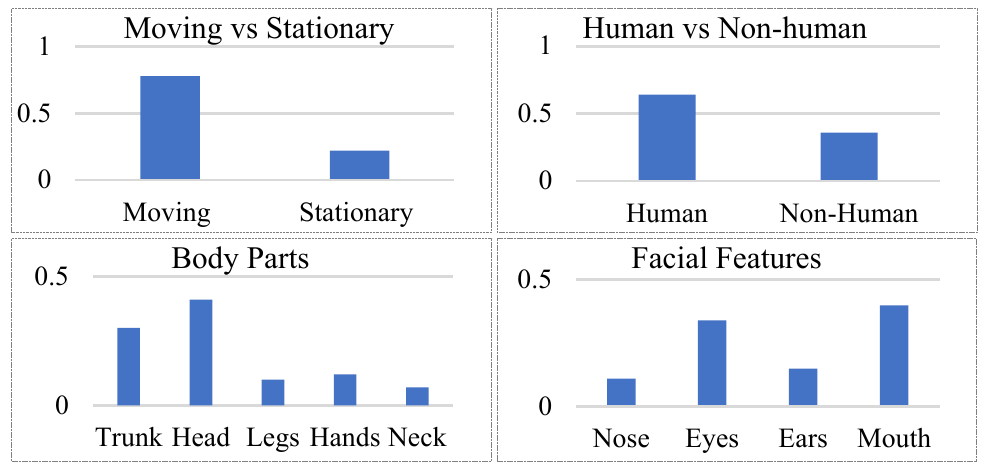}

   \caption{The objects of interests in visual attention. The y-axis denotes possibilities of objects drawing the human attention.}
   \label{fig:figure9}
\end{figure}

We performed a statistical experiment to investigate which kind of movie objects have higher possibility to draw participants’ attention. \cref{fig:figure9} shows the statistical results of our experiment. We start from the main category (moving objects vs. stationary objects) and end up with specific sub-categories (facial features). During the movie watching, most subjects pay their attention to the moving objects rather than the stationary objects. The difference is significant (0.783 vs 0.217), indicating that when the moving and stationary objects present at the same time, the information of stationary objects is rarely processed by the participants (21.7\% chance). This is probably because of the slow process of visual transduction, which has been suggested in \cite{berry1999anticipation} that visual stimulus evokes neural activity with a delay of 30-100 millisecond, therefore, the human brain may need to extrapolate the trajectory of a moving object in order to perceive its actual location. Among all the moving objects, participants are particularly interested in the human objects, and we are surprised to see that the human mouth has the highest possibility for drawing participants’ attention (\cref{fig:figure9}). The underlying reasons still need to be investigated, but there are some studies suggest that the attention on mouth usually related to language learning or the intention of creating more opportunities for communication \cite{barenholtz2016language}. Because all the participants are Germans while the movie was filmed using English, so it is likely that the participants were trying to process language information of movie characters by focusing on their mouths. 

\section{Conclusion}
In this work, we proposed a novel brain-inspired adversarial visual attention network (BI-AVAN) to characterize the human visual attention directly from functional brain activity. Our design of BI-AVAN model was inspired by the biased competition in the human visual system and can identify and locate the visual objects in a movie frame on which the human brain focuses. We evaluated the proposed BI-AVAN model with eye-tracking data and found that it achieved a high hit rate on both group-wise and individual-specific visual attentions. We also visualized the brain networks learned by the BI-AVAN model and discovered their strong correlations with the human visual attention. Finally, we studied the objects of interests in human visual attention statistically based on the proposed model. Overall, our BI-AVAN model contributes to the emerging field of leveraging the brain’s functional architecture to inspire and guide the model design in AI, e.g., deep neural network. In our future work, we will try to use even larger scale natural stimulus fMRI data to further improve and evaluate the BI-AVAN model.

\bibliographystyle{IEEEtran}
\bibliography{ref}

{\appendices
\section{Examples of Group-wise Visual Attention}
We provided 24 more examples of group-wise visual attention in \cref{fig:figureS4}.
\begin{figure*}[t]
  \centering
   \includegraphics[width=1.0\linewidth]{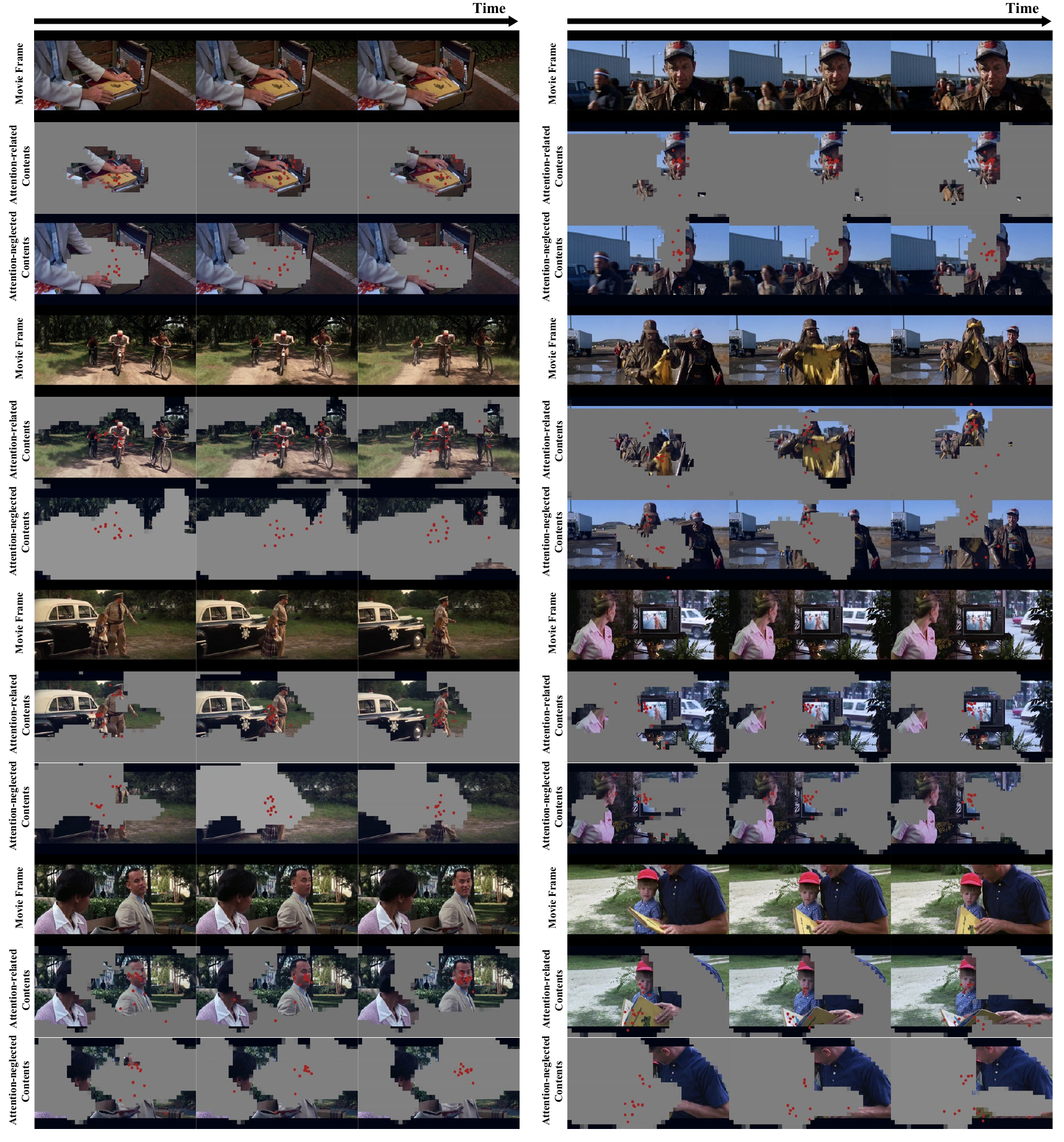}

   \caption{Examples of group-wise visual attention.}
   \label{fig:figureS4}
\end{figure*}

\section{Examples of Individual-specific Visual Attention}

We provided 12 more examples of individual-specific visual attention from 3 different subjects in \cref{fig:figureS5}.
\begin{figure*}[t]
  \centering
   \includegraphics[width=1.0\linewidth]{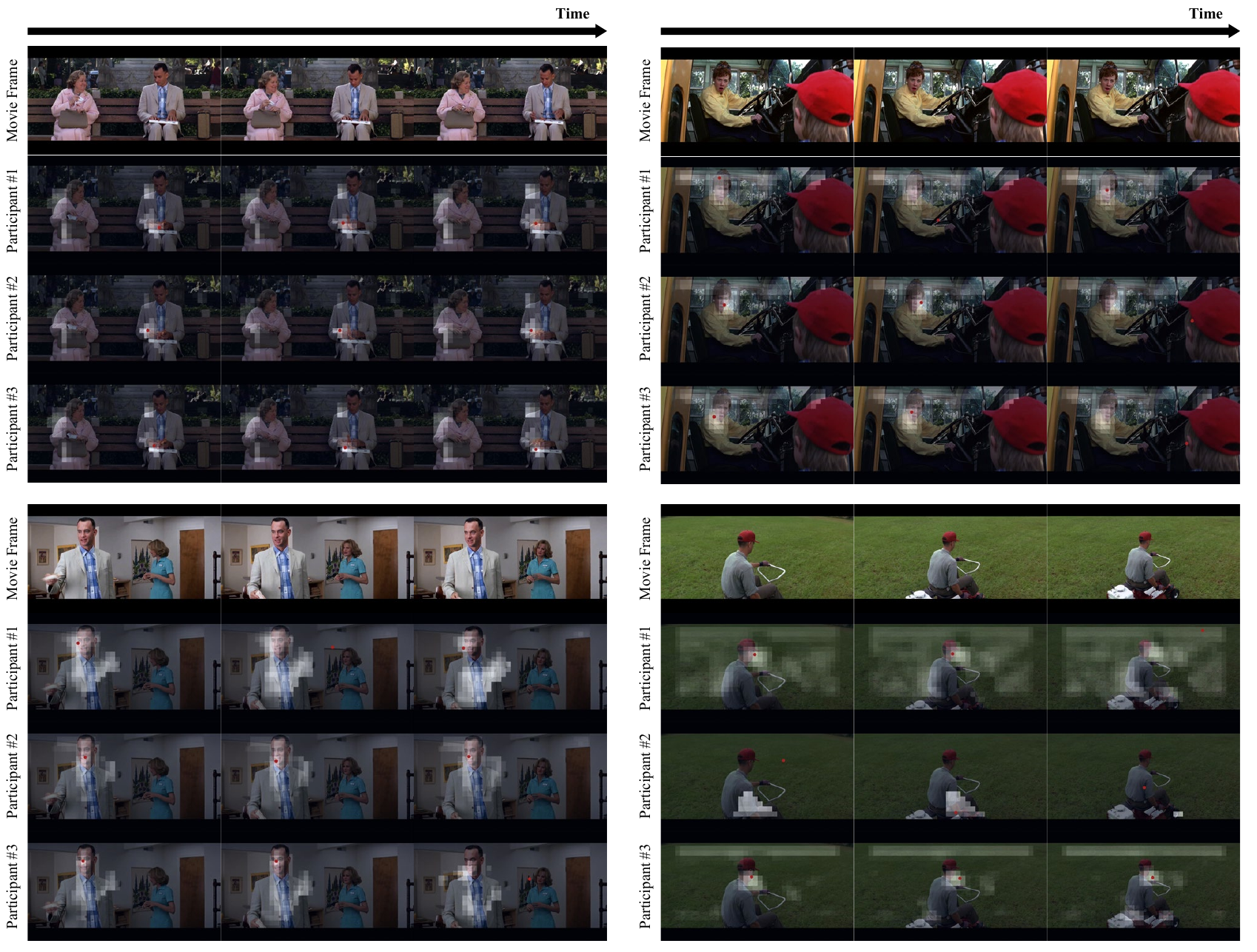}

   \caption{Examples of individual-specific visual attention from 3 different participants.}
   \label{fig:figureS5}
\end{figure*}

}

\end{document}